\documentclass[11pt]{article}
\usepackage{acl-hlt2011}
\usepackage{times}
\usepackage{graphicx}  
\usepackage{url}       
\usepackage{latexsym}
\usepackage{amsmath}
\usepackage{multirow}

\long\def\/*#1*/{}

\begin{document}

\title{PR2: A Language Independent Unsupervised Tool for Personality Recognition from Text}
\author{Fabio Celli \\ CLIC-CIMeC, University of Trento \\ \texttt{fabio.celli@unitn.it}\\
\And Massimo Poesio\\University of Essex\\ \texttt{poesio@essex.ac.uk}}
\maketitle
\begin{abstract}
We present PR2, a personality recognition system available online, that performs instance-based classification of Big5 personality types from unstructured text, using language-independent features. It has been tested on English and Italian, achieving performances up to f=.68.\\
\end{abstract}

\section{Introduction and Background}
Personality is an affect processing system that describes persistent human behavioural responses to broad classes of environmental stimuli (Adelstein et al. 2011)
. It has been formalized in various ways, such as the Myers-Briggs type indicator (Briggs \& Myers 1980)
, that defines four personality types, and the Big5 (Costa \& MacCrae 1992)
, that defines five bipolar traits, is widely used in the scientific community, and can be assessed by means of different questionnaires. 
The Big5 traits are: extroversion, emotional stability/neuroticism, agreeableness, conscientiousness and openness to experience. 
Written text, as well as speech, conveys a lot of information about author's personality (Mairesse et al. 2007)
. Personality Recognition from Text is a NLP task, partially connected to stylometry (Luyckx \& Daelemans 2008)
, that consists in the automatic classification of authors' personality traits from linguistic features. Useful applications of this task range from deception detection (Fornaciari et al. 2013) 
to social network analysis (Celli \& Polonio 2013), and potentially many NLP tasks, such as opinion mining. In order to find new applications of personality in language, we present PR2, a personality recognition tool for NLP available online\footnote{http://clic.cimec.unitn.it/fabio/pr2demo.php}.\\ 
\indent The paper is structured as follows: in section \ref{prev} we introduce previous work and the problems we see in personality recognition from text. In section \ref{sys} and \ref{exp} we describe our tool for personality recognition and the experiments to test its performance. In section \ref{end} we draw some conclusions.

\section{Previous Work, Problems and Perspectives}\label{prev}
After the first pioneering works in personality recogniton in blogs and offline texts (Oberlander \& Nowson 2006;
Mairesse et al 2007
), recently there has been an increasing interest in the extraction of personality from social networks and in languages different from English (Kermanidis 2012; 
Bai et al 2012; 
Quercia et al. 2011; 
Golbeck et al 2011
). \\
\indent Almost all the approaches to personality recognition from text are supervised. The main problems with this approach are the limitations in data annotation and language dependency. Data labeled with personality types are usually costly and time consuming to collect. In addition, when models are trained on a specific domain or language, are not very effective if used on different domains. The language problem also concerns the language dependency of the resources, such as LIWC (Tausczik \& Pennebaker 2010) 
and MRC (Coltheart 1981). 
These are clear limitations in the exploitation of personality recognition from text in other NLP tasks.\\
\indent Many scholars working in the field of personality recognition from text reported correlations between linguistic cues and personality traits (Mairesse et al 2007; Iacobelli et al 2011; Quercia et al 2011; Golbeck et al. 2011) 
that can be exploited as models for unsupervised classification (Celli 2012). 
Unsupervised personality recognition has some advantages: they require very few labeled data (mainly for validation) and are potentially language-independent. In the next section we describe our system for Unsupervised personality recognition from text.

\section{PR2: System description, Feature Set and Parameters}\label{sys}
PR2 is a personality recognition tool written in Perl and available online as a demo. It performs instance-based classification of Big5 personality types in an unsupervised way, using language-independent features (see table \ref{feat.c}). The system takes as input unlabeled text and  authors in a tab separated format, with authors in the first column and text in the second one. Examples of the input format are provided on the website. Big5 personality labels are formalized as 5-characters strings, each one representing one trait of the Big5. Each character in the string can take 3 possible values: positive (y), negative (n) and omitted/balanced (o). For example ``ynoon'' stands for an extrovert neurotic and not open mindend person. \\
\indent As initial feature set, the system exploits language-independent features extracted from LIWC 
and MRC
, whose correlations to personality are reported in Mairesse et al. 2007
. 
\begin{table}[ht]
\centering \small
\begin{tabular}{lccccc}
\hline
feature & ext. & emo. & agr. & con. & ope.\\
\hline
ap & -.08** & -.04 & -.01 & -.04 & -10** \\
em &  -.00 & -.05* & .06** & .00 & -.03 \\
nb &  -.03 & .05* & -.03 & -.02 & -.06** \\
pa &  -.06** & .03 & -.04* & -.01 & .10** \\
qm &  -.06** & -.05* & -.04 & -.06** & .08** \\
qt &  -.05* & -.02 & -.01 & -.03 & .09** \\
tt & -.05** & .10** & -.04* & -.05* & .09**\\
wf & .05* & -.06** & .03* & .06** & .05**\\
\hline
\end{tabular}
\caption{\small{Feature/correlations set, adapted from Mairesse et Al. 2007.  * = \textit{p} smaller than .05 (weak correlation), ** = \textit{p} smaller than .01 (strong correlation).} \label{feat.c}}
\smallskip
\end{table} 
The features are: punctuation (ap); question marks (qm); quotes (qt); exclamation marks (em); numbers (nb); parentheses (pa); repetition ratio (tt), word frequency (wf).\\
\indent The pipeline of the personality recognition system has three phases. In the preprocessing phase the system filters repeated texts and sorts the authors that have more than one text. Then it samples 20\% of the input unlabeled data, assigns personality labels according to the correlations and computes the distribution of each feature of the correlation set and their firing rate. 
\begin{figure}[h]
\begin{center}
\includegraphics[width=7.2cm]{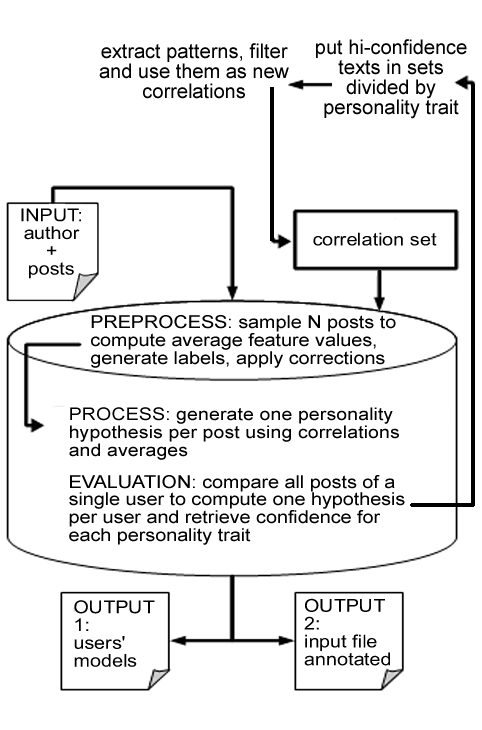}
\caption{\small{System pipeline.}} \label{spipe}
\end{center}
\end{figure}
In the processing phase, the system generates one personality hypothesis for each text in the dataset, mapping the features in the correlation set to specific personality trait poles, according to the correlations. Instances are compared to the average of the population sampled during the preprocessing phase and filtered accordingly: only feature values above the average are mapped to personality traits. For example a text containing more punctuation than average will fire negative correlations with extraversion and openness to experience (see table \ref{feat.c}). The system keeps track of the firing rate of each single feature/correlation and computes personality scores for each trait, mapping positive scores into ``y'', negative scores into ``n'' and zeros into ``o'' labels. In this phase the system computes also per-trait confidence, defined as the coverage of the majority label over all the labels generated for the author's texts. $ \textit{conf}=\frac{\textit{m}}{\textit{T}}$ where \textit{m} is the count of the majority label, and \textit{T} is the count of the author's texts. In the evaluation phase the system compares all the personality hypotheses generated for each single text of each author and retrieves one generalized hypothesis per author by computing the majority class for each trait. In the evaluation phase the system computes average confidence and variability. Average Confidence is derived from per-trait confidence scores and gives a measure of the robustness of the personality hypothesis. Variability (var) gives information about how much one author tends to write expressing the same personality traits in all the texts. It is defined as $ var=\frac{\textit{avg conf}}{\textit{T}}\label{v} $ where \textit{avg conf} is the confidence averaged over the five traits and \textit{T} is the count of all author's texts. The outputs of the system are: 1) list of authors with personality labels, confidence, number of texts and variability; 2) input text annotated with personality labels in a tab-separated format.\\
\indent The system provides the following optional parameters:
\textbf{weigthed correlations (w)}. When the parameter is activated, high feature firing rates, computed on the fly during the processing phase, decrease the personality score associated to that feature. This parameter boosts the infomation provided by low-frequency features.  
\textbf{variable hypothesis average (v)}. If this parameter is activated, the average distribution of each feature is recomputed on the fly during the processing phase. This allows to fit the data at hand but increases the error rate on the first instances processed. 
\textbf{hypotheses normalization (n)}. If this parameter is activated, the system normalizes hypotheses scores during the processing phase, allowing a better comparison of the authors. If paired with variable hypothesis average, these parameters force binary classification (no ``o'' labels), boosting recall. 
\textbf{weak traits correction (r)}. When the distribution of specific features is paricularly skewed, the system might generate labels just for one class. If this parameter is activated, the system detects skewed distributions for particular traits during the preprocessing phase and randomizes the personality score associated to those traits. This parameter can be useful to prevent errors with very small datasets. 
\textbf{pattern extraction (t)}. If this parameter is activated, the system automatically extracts new patterns from the data at hand, filtering out labels with low confidence, associates them to personality traits, and uses them as new correlations between patterns and personality traits. 

\section{Experiments: Testing the System}\label{exp}
We tested the system on two datasets different in size, domain and language: Essays (Mairesse et al 2007
), a large collection of essays written in English, and PersFB (Celli \& Polonio 2013
), a small dataset of Facebook posts in Italian. We split Essays by lines in order to have more or less the same length per each texts in the two datasets (19 words per text in Essays, and 12 words per text in PersFB). As test sets, we used a sample of 250 authors for Essays and about 25 authors for PersFB. Patterns (parameter t) were extracted from larger sets: 2000 authors for Essays, and about 1000 authors for PersFB.
\begin{table}[ht]
\centering \small
\begin{tabular}{lllll}
\hline
dataset & par & p & r & f\\
\hline
mbl-fb-it & - & .437 & 1 & .608 \\
fb-it & n  &  .477  & .855  & .612\\
fb-it & nw  &  .492  & .87  & .629\\ 
fb-it & nr  &  .478  & .86  & .614\\
fb-it & nv  &  .472  & 1  & .641\\
fb-it & nvr  &  .493  & 1  & .661\\
fb-it & tnvr & .534 & 1 & \textbf{.686}\\
\hline
mbl-es-en & - & .487 & 1 & .655 \\
es-en & n  &  .544  & .861  & .667\\
es-en & nw  &  .525  & .855  & .651\\
es-en & nr  &  .549  & .908  & .684\\
es-en & nv  &  .537  & 1  & \textbf{.699}\\
es-en & nvr  &  .536  & 1  & .698\\
es-en & tnvr & .523 & 1 & \textbf{.686}\\
\hline
\end{tabular}
\caption{\small{Average precision, recall and f-measure for different datasets in a 2-way classification task. 2-tailed test. Results are averaged over the five personality traits.}\label{parx1}}
\end{table}
 Since each personality trait is bipolar, we decided to run the tests, considering as true positives the correct predictions for both poles, as false positives the wrong predictions and as false negatives the missing predictions. The majority baseline is the mean of the predictions using all ``y'' labels and all ``n'' labels. Results, reported in table \ref{parx1}, show that similar performances can be obtained on very different data types. 

\section{Conclusions and Future Work}\label{end}
We presented PR2: a NLP tool, available online, for language-independent unsupervised personality recogniton from unlabeled text. Unlike supervised systems, that require large labeled datasets for training, this tool requires just a small labeled dataset for the validation of the annotation.\\
\indent In the future, we would like to improve the precision of the system and to provide a scorer for validation. 









\end{document}